\documentclass[10pt, a4paper]{article}

\usepackage{lrec-coling2024}

\usepackage{listings}
\usepackage[table]{xcolor}
\usepackage{natbib}
\usepackage{multibib}
\makeatletter
\def\@mb@citenamelist{cite,citep,citet,citealp,citealt,citepalias,citetalias}
\makeatother
\newcites{languageresource}{~}

\usepackage{graphicx}
\usepackage{tabularx}
\usepackage{hhline}
\usepackage{multirow}
\usepackage{ifthen}

\usepackage{soul}

\usepackage{authblk}    
\usepackage{adjustbox}
\usepackage{amssymb,amsthm,amsmath}
\usepackage[caption=false]{subfig}
\usepackage{booktabs}
\usepackage{tikz}
    \usetikzlibrary{tikzmark}
    \usetikzlibrary{calc} 
    \usetikzlibrary{shapes,shadows,arrows,positioning} 
    \usetikzlibrary{decorations.pathreplacing,decorations.pathmorphing} 

\usepackage{pgfplots}
    \usepgfplotslibrary{colorbrewer} 
    \usepgfplotslibrary{groupplots} 
    \usepgfplotslibrary{fillbetween} 
    \pgfplotsset{compat=newest}
\usepackage{pgfplotstable} 

\pgfplotsset{    
        x label style={at={(axis description cs:0.5,-0.15)},anchor=north},
        ylabel style={at={(axis description cs:-0.1,.5)},anchor=south},
        ylabel near ticks, 
        ylabel shift = -0.5em, 
}

\pgfplotsset{every axis/.append style={
                    legend style={font=\tiny,line width=.5pt,mark size=.6pt},
                    }}
\pgfplotsset{
        axis on top,
        xtick align=inside,
}

\pgfplotsset{
  log x ticks with fixed point/.style={
      xticklabel={
        \pgfkeys{/pgf/fpu=true}
        \pgfmathparse{exp(\tick)}%
        \pgfmathprintnumber[fixed relative, precision=3]{\pgfmathresult}
        \pgfkeys{/pgf/fpu=false}
      }
  },
  log y ticks with fixed point/.style={
      yticklabel={
        \pgfkeys{/pgf/fpu=true}
        \pgfmathparse{exp(\tick)}%
        \pgfmathprintnumber[fixed relative, precision=3]{\pgfmathresult}
        \pgfkeys{/pgf/fpu=false}
      }
  }
}

\tikzset{
>=latex,
punkt/.style={
       rectangle,
       rounded corners,
       draw=black, very thick,
       text width=6.5em,
       minimum height=2em,
       text centered},
      punkt2/.style={
       rectangle,
       rounded corners,
       draw=white, very thick,
       text width=6.5em,
       minimum height=2em,
       text centered}, 
}

\usepackage{titlesec}
\titleformat{\section}{\normalfont\large\bfseries\center}{\thesection.}{1em}{}
\titleformat{\subsection}{\normalfont\SmallTitleFont\bfseries\raggedright}{\thesubsection.}{1em}{}
\titleformat{\subsubsection}{\normalfont\normalsize\bfseries\raggedright}{\thesubsubsection.}{1em}{}
\renewcommand\thesection{\arabic{section}}
\renewcommand\thesubsection{\thesection.\arabic{subsection}}
\renewcommand\thesubsubsection{\thesubsection.\arabic{subsubsection}}

\usepackage{xcolor}
\usepackage{hyperref}
 \definecolor{darkblue}{rgb}{0, 0, 0.5}
  \hypersetup{colorlinks=true, citecolor=darkblue, linkcolor=darkblue, urlcolor=darkblue}

\usepackage{xstring}

\usepackage{color}

\title{Code-Mixed Probes Show How Pre-Trained Models Generalise On Code-Switched Text 
\vspace*{.5\baselineskip} 
}

\name{Frances A. Laureano De Leon\textsuperscript{\textasteriskcentered}, Harish Tayyar Madabushi\textsuperscript{\textdagger}, Mark Lee\textsuperscript{\textasteriskcentered}} 

\address{\textsuperscript{\textasteriskcentered}University of Birmingham, \textsuperscript{\textdagger}University of Bath \\
         \textsuperscript{\textasteriskcentered}School of Computer Science, Birmingham, UK\\
         \textsuperscript{\textdagger}Department of Computer Science, Bath, UK \\
         fxl846@cs.bham.ac.uk, htm43@bath.ac.uk, m.g.lee@bham.ac.uk\\}

\abstract{
Code-switching is a prevalent linguistic phenomenon in which multilingual individuals seamlessly alternate between languages. Despite its widespread use online and recent research trends in this area, research in code-switching presents unique challenges, primarily stemming from the scarcity of labelled data and available resources. In this study, we investigate how pre-trained Language Models handle code-switched text in three dimensions:
a) the ability of PLMs to detect code-switched text, b) variations in the structural information that PLMs utilise to capture code-switched text, and c) the consistency of semantic information representation in code-switched text.
To conduct a systematic and controlled evaluation of the language models in question, we create a novel dataset of well-formed naturalistic code-switched text along with parallel translations into the source languages. Our findings reveal that pre-trained language models are effective in generalising to code-switched text, shedding light on the abilities of these models to generalise representations to CS corpora. We release all our code and data, including the novel corpus, at~\url{https://github.com/francesita/code-mixed-probes}. 
 \\ \newline   
}

\begin{document}

\maketitleabstract

\section{Introduction}
\label{intro}

Code-switching (CS) is the phenomenon in which multilinguals effortlessly alternate between languages in the same conversation or piece of writing~\cite{Joshi1982, Dogruoz2021}.
CS arises in multilingual communities over the world, such as the United States, Latin America, and India, and gives way to the emergence of mixed 'languages' such as Hinglish (Hindi-English mix) and Spanglish (Spanish-English mix).
The recent adoption of Pre-trained Language Models (PLMs) has been driven in part by their ability to gain a significant amount of linguistic information~\cite{Clark2019, Tenney2019b} and world knowledge~\cite{Petroni2019} based purely on the pre-training. 
A significant question is how much information PLMs can gather about the meaning of words from being trained on text alone~\cite{Bender2020}.
CS data is especially useful in helping to answer this question due to the multilingual nature of CS text.
The presence of multiple languages in the text will prevent the model from relying on spurious statistical correlations when generating meaning because (1) the presence of multilingual text will encourage the model to learn meaning across languages, and (2) the model needs to learn the context of each language switch, which may prevent it from using simple patterns more easily found in monolingual text due to language-specific patterns.  
Hence, the semantic representations of CS data provide us a way of exploring the true extent to which models are able to capture and generalise meaning.

Despite the potential significance of exploring CS data in evaluating PLMs, research in this area is challenging, not least due to the lack of labelled data and resources~\cite{Santy2021, Lince2020}.
As such, in this work, we focus on how PLMs interact with and encode code-switched text.
To ensure a balanced evaluation of models, we look at both real and synthetic CS text and focus exclusively on Spanglish (Spanish-English).
We choose Spanglish for 3 reasons: (1) Spanish and English share a script, 
(2) English words share many Spanish cognates. This overlap is due to a portion of English vocabulary having Latin roots, and Spanish having originated from Colloquial Latin,~\cite{Nagy1993}, and
(3) although there are differences in Spanish and English grammar, such as word-order, gender and number, there also are overlaps in the structure of both languages~\cite{Rivera2019}.
These similarities ensure that our evaluation of model capabilities are not confounded by other aspects of language use, such as, for example, difference in script as in the case of Hinglish.
We utilise synthetic data in our experiments to investigate whether the presence of mixed language text alone leads to satisfactory probe performance, or whether the use of naturalistic CS examples significantly impact experimental results. 

To evaluate the extent to which PLMs can identify and encode the correct representations of CS text, we focus our efforts on three different dimensions: 
a) the ability of PLMs to \emph{detect} CS text, 
b) possible variations in the grammatical structural that PLMs are able to capture from CS text, and
c) consistency of the meaning representations of CS text when compared with monolingual text. Our experimental results along these lines indicate that PLMs have the potential to capture all three of these dimensions with reasonable exactness.
While further experiments are required in this regard, our findings seem to indicate that PLMs are surprisingly good at generalising across CS text, which could shed light on the potential of PLMs to capture some generalisations pertaining to language use. 

To this end, we perform multiple experiments, largely using probes, to evaluate each of these different dimensions.
We probe popular pre-trained models: mBERT~\cite{Devlin2019}, and XLM-RoBERTa base (XLM-R-base)~\cite{Conneau2018}, which consist of 12 layers and 768 dimensions, and XLM-RoBERTa large (XLM-R-large)~\cite{Conneau2018} with 16 layers and 1024 dimensions.
We begin by exploring relevant literature associated with CS text, probing, graph edit distance and existing datasets in section~\ref{sec:lit-review} before then describing the construction of a well-balanced corpus of CS text that we create due to existing limitations of availability of such data in section~\ref{sec:datasets}. We then detail our experiments in each of these directions in Sections~\ref{sec:detection},~\ref{sec:syntax}, and~\ref{sec:semantics}, where we present our methods and results. We follow by a discussion of these results in Section~\ref{sec:discussion} and provide a summary of our findings and suggestions for future work in Section~\ref{sec:conclusion}. 

\subsection{Contributions}

Given the importance of research in CS, this work makes the following contributions: 
We create the first curated dataset of well-formed, naturalistic instances of Spanglish CS data with translations for both source languages to allow for a precise evaluation of grammatical structure and sentence meaning.
We perform extensive experiments to determine the extent to which PLMs can detect CS text and capture both the structure and meaning associated with CS text.
Additionally, we extend our manually curated dataset with synthetic data to allow for ablation studies which include various controls such as the mix of languages in CS text.
We provide a template for future experimental verification of linguistic theories pertaining to CS based on the usage-based principle of language acquisition.

\section{Related Work}
\label{sec:lit-review}
In this section we discuss related research associated with CS data, probes and methods of comparing language structure. 

\subsection{Code-switching and data generation}
As previously mentioned, CS has become more available thanks to the rise of social media and multilingual users~\cite{Winata2023}.
To facilitate research, datasets and evaluation benchmarks, such as LinCE, have been created in an effort to have a centralised evaluation platform for code-switching,~\cite{Lince2020}.
LinCE combines ten corpora covering four different code-switched language pairs and four tasks.
 \citet{Khanuja2020}  also provide a generalised CS benchmark that is inspired by GLUE known as GLUECoS. 
 Despite these efforts, research in the domain remains challenging due to reasons mentioned in Section~\ref{intro}. Our work in introducing a novel dataset is aimed at addressing this shortcoming. 
This has led to growing research in synthetic data generation for CS text, which motivates us to expand our manually curated dataset with synthetic data, see Section~\ref{sec:datasets}. 
Some of the techniques employed to generate synthetic CS data in previous works include: 
(1) Identification and replacement of noun-phrases in monolingual sentences with the translation of that phrase in the other language pair to be studied~\cite{Salaam2022}.
(2) Generation of CS text containing randomly selected languages to create a CS example containing switching in multiple languages~\cite{Krishnan2021}.  
This data was used to create a model referred to as "modified mBERT", which is trained on synthetic and real code-switched data and then tested on NLI in Hinglish.
(3) The use of models trained on CS text generation~\cite{Winata2019, Rizvi2021}. We use the first two of these methods to augment our dataset with synthetic data.

Many of the synthetic data generation methods are inspired or driven by CS grammar theories developed in the field of linguistics~\citep{Bullock2009, Sebba2012}.
There are two CS theories that take precedence within NLP, the Equivalence Constraint theory (EC), in which language switches occur when the surface structures of languages align~\cite{Poplack1980} and Matrix Language Frame (MLF) model, in which one language is dominant and determines the syntax of a CS phrase~\cite{Joshi1982, Myers1993}. 
Although there are other grammar theories explaining CS, these are the most used in NLP for the creation of synthetic data.
EC theory states that alternations between languages occur when the surface structures of the languages align, therefore the grammar rules of both languages are obeyed.  
Broadly, the MLF theory holds that in CS sentences, there is a matrix language and an embedded language.
The matrix language is that which provides the grammatical structure that accommodates words or phrases from another language~\cite{Dogruoz2021}. 
Our exploration of the manner in which the syntactic information pertaining to CS text is encoded in PLMs is driven by this theoretical work in linguistics. Given that there are competing theories explaining the use of CS languages, our experiments are designed to evaluate if the grammatical structure of CS data extracted by PLMs is independent of either source languages, see Section~\ref{sec:syntax}. 

\subsection{Probes}
In this section, we introduce literature related to probes, which we use extensively in this work. 
Probes, also known as auxiliary or diagnostic classifiers~~\citep{Adi2017}, have been developed to investigate linguistic properties  encoded in text representations ~\citep{Tenney2019}.
They have been used  for extrinsic exploration, in which a machine learning model is used to determine whether a linguistic structure in present in representations through performance on a task such as named entity recognition~\cite{Hennigen2020} and intrinsic exploration, which looks to evaluate representations on benchmarks regarding the relationship between words or sequences~\citep{Linspector2020}. 
Probes have been used for a number of years and have been largely used to analyse morphological, semantic and syntactic language properties~\citep{Dalvi2019}.
Probes are necessarily simply classifiers used to predict a property of some input text~\citep{Adi2017} based on the representations generated by a model, and often consist of  a linear layer, or multilayer perceptron on top of frozen representations.
Generally,  the word or sentence representations studied are frozen, in order to prevent further training of the representations and are used as the embedding inputs for the probe classifier.
As the representations are frozen,  if a probe classifier learns to predict the property it was trained on, it is an indication that there is a linear mapping between the internal representations of the model and the required output and so an indication of that property being embedded within the model.

Works relevant to us in the field of probing is the syntactic structural probe by~\citet{Hewitt2019}, in which they find that syntax trees are embedded in a deep models' representations.
This work is expanded on  by~\citet{Chi2020}  , who use the structural probe and find that syntactic features overlap between languages, which agrees with universal dependencies' taxonomy in mBERT.  
\citet{Chi2020} also find that the structural probe most effectively recovers tree structure from the 7th or 8th mBERT layer, and that a maximum rank beyond 64 or 128 gives no further gains.
\citet{Tenney2019} introduce a framework they call "edge probing", which provides a uniform architecture across tasks.
They use the edge-probing technique to do layer-wise explorations of the BERT model, in which they find that basic syntactic information appears earlier in the network, and high-level semantic information appears at the higher layers~\cite{Tenney2019b}.

Prior work probing the syntactic structure of CS text has been limited:
~\citet{Pires2019}, as part of their study, use a POS dataset to probe mBERT on code-switched text.
A more detailed probe study was done by~\citet{Santy2021}, in which synthetically generated and real code-mixed data are used to probe mBERT.
They compare the probe results for different tasks, such as POS, NER, LID  to the fine-tuned version of the model trained for that task.
They find that using synthetically generated data in certain tasks yield lower results than using naturally occurring code-switched data.

\subsection{Syntax and Graph Edit Distance}
An important aspect of our work is in evaluating the similarity of syntactic structure extracted by probes. In this section, we review relevant work pertaining to the comparison of such structures. 
Graph Edit Distance (GED) is a metric commonly used for structural pattern recognition and analysis of graphs~\cite{Gao2010}.
GED is used on dependency parses, where the parses are represented by unordered directed trees in order to filter out sentence pairs that cannot be compared syntactically~\cite{Kroon2019}.
\citet{Kroon2019} utilise this method for the massive automatic syntactic comparison of languages.
Unordered graphs make it so that the GED algorithm is more robust between different languages, which is a reason they find GED to be a good technique for syntax comparison between different languages.
They favour the use of parallel corpora for automatic comparisons because it facilitates finding the contexts in which differences in syntax occur~\cite{Kroon2019}. 

\subsection{Existing CS Datasets}
In this section, we discuss existing datasets for Spanglish text.
Although CS datasets are generally scarce, Spanglish is a popular language pair, in which some CS data can be found~\cite{Winata2023}.
Many of the publicly available datasets are from shared tasks, such as CALCS workshops~\cite{Winata2023}.
Some of the most used data in research include language identification (LID) data from a shared task in 2016 by~\citet{Molina2016}, SentiMix 2020 sentiment analysis dataset by~\citet{Patwa2020},  and datasets for part-of-speech classification (POS)~\cite{Alghamdi2016} and named entity recognition (NER)~\cite{Aguilar2018} created for shared tasks.
All these datasets consist of tweets, apart from the POS dataset, which is derived from the Miami Bangor Corpus, and consists of bilingual and CS conversations from four speakers.
This dataset is annotated with Universal POS tags by~\citet{Soto2017}.
All the aforementioned datasets contain LID labels.
There is also a machine translation dataset for Spanglish available that was created for CALCS 2021, but this dataset does not contain parallel translations~\cite{Chen2021}.
All of these datasets are available on the LinCE website~\footnote{\url{https://ritual.uh.edu/lince/datasets}}.

\begin{table*}[htb]
\centering
\resizebox{\linewidth}{!}{%
\begin{tabular}{|l|l|l|l|l|l|l|}
\hline
\multicolumn{1}{|l|}{\textbf{dataset}}                            & \multicolumn{1}{l|}{\textbf{\# tokens}} & \multicolumn{1}{l|}{\textbf{en}} & \multicolumn{1}{l|}{\textbf{es}} & \multicolumn{1}{l|}{\textbf{other}} & \multicolumn{1}{l|}{\textbf{ne}} & \multicolumn{1}{l|}{\textbf{unk}} \\ \hhline{|=|=|=|=|=|=|=|}
Real CS data                                                      & 4302                                    & 2174 (50.53\%)                   & 1397 (32.47\%)                     & 657 (15.27\%)                       & 73 (1.69\%)                      & 1 (0.02\%)                        \\
Random CS data                                                    & 4649                                    & 2039 (43.85\%)                     & 1867 (40.16\%)                   & 662 (14.24\%)                       & 80 (1.72\%)                      & 1 (0.02\%)                        \\
En noun phrase  synthetic CS data& 4233                                    & 1338 (31.61\%)                     & 2145 (50.67\%)                     & 673 (15.90\%)                         & 76(1.79\%)                       & 1(0.02\%)                         \\
Es noun phrase synthetic CS data& 4640                                    & 2411 (51.96\%)                   & 1456 (31.38\%)                   & 657 (14.16\%)                       & 116 (2.5\%)                      & 0                                \\
\hline
\end{tabular}
}
\caption{Created CS datasets: en stands for English, es for Spanish, other for largely punctuation, ne for named-entities and unk for unknown.}
\label{tab:cs_dataset_stats}
\end{table*}

As far as we know, there is no available naturalistic Spanglish dataset that includes translations for BOTH source languages.
Such a dataset is essential for conducting a systematic and controlled evaluation of the PLMs under investigation. 
Hence, we create such a dataset, which stands as one of our contributions to the research.

\section{Dataset Creation}
\label{sec:datasets}
Due to the absence of naturalistic datasets containing Spanglish data with associated parallel Spanish and English, we construct a novel dataset to address this shortcoming, see table~\ref{tab:cs_dataset_stats}. 

\subsection{CS Data Collection}
We collect CS data from X, previously known as Twitter, using the techniques described in~\citet{Laureano2020}.
This method uses a keyword file that contain the most commonly used words in one of the language pairs (i.e. Spanish).
We use the top 100 most frequent words used in Spanish according to the Dictionary of the Royal Spanish Academy~\footnote{\url{https://corpus.rae.es/frec/5000_formas.TXT}}, and filter out words that contain 4 letter or less to prevent overlap with other languages and remove articles and pronouns.
To ensure a CS output, the search query should specify the other language pair to be studied (i.e. English).
We select a random subset of posts from the collected tweets to be part of the CS dataset that we use to test our probes. 
A person fluent in Spanish and English helped check this subset of tweets for real occurrences of CS in Spanglish and to discard any unusable or incoherent posts. 
These CS posts were then translated into Spanish and English by a speaker of both languages, with the aid of Google Translate API~\footnote{\url{https://pypi.org/project/googletrans/}}.
We ultimately obtain a total of 316 posts after quality checks and translations.
A subset of this collected data is used as part of our syntax and semantic experiments. 
Specifically, we choose examples containing intra-sentential CS, the type of CS in which language alternations happens within a sentence.
Intra-sentential instances of CS are essential to observe with confidence the interaction between two grammars~\cite{Joshi1982}, and are, therefore, key for the syntax experiments, see Section~\ref{sec:syntax}.
We gather 254 intra-sentential examples to use for syntax experiments, and refer to these as~\textbf{\textit{r-CS}} to denote that they are real instances of CS.
These 254 examples were chosen on the basis of whether they were instances of intra-sentential CS.
We remove hashtags, links from these examples.
The examples were also re-written by bilinguals in Spanish and English, in order to create well-formed sentences, which is challenging to find in social media.
View table~\ref{tab:dataset-examples} for examples of the original posts, the edited CS text and translations into source languages.

{
\renewcommand{\arraystretch}{1.3}%
\begin{table*}[]
\centering
\resizebox{\textwidth}{!}{%
\begin{tabular}{|l|l|l|l|}
\hline
\textbf{Original post} & \textbf{Edited post} & \textbf{Spanish translation} & \textbf{English translation} \\ 
\hhline{|=|=|=|=|}
\begin{tabular}[c]{@{}l@{}}siempre me dicen que no sea tan inseguro,\\ i'M tRyInG mY bESssT. {\includegraphics[height=1em]{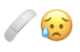}} \end{tabular} & \begin{tabular}[c]{@{}l@{}}Siempre me dicen que no sea tan \\ inseguro,  I’m trying my best. \end{tabular} & \begin{tabular}[c]{@{}l@{}}Siempre me dicen que no sea tan\\ inseguro, Estoy tratando. \end{tabular} & \begin{tabular}[c]{@{}l@{}} They always tell me not to be so \\ insecure, I'm trying my best.\end{tabular} \\ \hline
\begin{tabular}[c]{@{}l@{}}NO HABIA VISTO QUE HE WAS ALMOST \\SHIRTLESS {\includegraphics[height=1em]{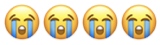}} \\ \url{https://t.co/d5tKtpzPMw} \end{tabular}& \begin{tabular}[c]{@{}l@{}} No había visto que he was almost \\ shirtless.\end{tabular} & \begin{tabular}[c]{@{}l@{}} No había visto que estaba casi sin \\ camisa.\end{tabular} & \begin{tabular}[c]{@{}l@{}}I hadn't seen that he was almost \\ shirtless.\end{tabular} \\ \hline
\begin{tabular}[c]{@{}l@{}} @rcknatsu first u gotta inhalar el aire \\ hacia los pulmones \end{tabular}& \begin{tabular}[c]{@{}l@{}}First you gotta inhalar el aire hacia \\ los pulmones.\end{tabular} & \begin{tabular}[c]{@{}l@{}}Primero tienes que inhalar el aire hacia \\ los pulmones.\end{tabular} & \begin{tabular}[c]{@{}l@{}} First you gotta inhale the air into \\ your lungs.\end{tabular} \\
\hline
\end{tabular}%
 }
\caption{Examples of original posts collected from X, and minimal editions and translations.}
\label{tab:dataset-examples}
\end{table*}
}

\subsection{CS data generation}

We utilise the parallel translations of the CS data we collect (\textbf{\textit{r-CS}}) to generate synthetic CS data using two different techniques found in literature, random replacement of a token in either of the language pairs~\cite{Krishnan2021}, and the noun-phrase replacement technique~\cite{Salaam2022}. 
The source data to generate the synthetic examples come from the English and Spanish translations of the \textbf{\textit{r-CS}} dataset.
For the random generation method, we tokenize the examples, and randomly choose whether that token should be translated or not. 
If translated, that token is replaced by the translation. 
For the noun-phrase synthetic dataset, we follow a 3-step process as described in~\citet{Salaam2022}.
(1) Noun-phrase identification, we use the spaCy library to do this~\footnote{\url{https://spacy.io/}},
(2) translate the noun-phrase into the desired language,
(3) replace the correct span with the translated noun-phrase.
We generate noun-phrase synthetic examples with both Spanish monolingual and English monolingual translations of our \textit{\textbf{r-CS}}, to ensure we have examples with majority Spanish (\textit{\textbf{NP-CS-es}}) and majority English tokens (\textit{\textbf{NP-CS-en}}).

\begin{table*}[]
\resizebox{\linewidth}{!}{%
\begin{tabular}{|llll|}
\hline
\textbf{Category} & \textbf{Experiment} & \textbf{Explanation} & \textbf{Aim} \\ \hhline{|====|}
Detection & Sentence Classification & \begin{tabular}[c]{@{}l@{}}Train probe classifiers for each PLM and layer to detect \\ whether a sentence is monolingual  or code-switched.\end{tabular} & \begin{tabular}[c]{@{}l@{}} Find if models can distinguish \\ between monolingual and CS sequences \end{tabular}\\ \hline
Detection & Language Identification (LID) & \begin{tabular}[c]{@{}l@{}}Train probe classifiers for each PLM per layer to learn \\the language ID of  tokens of CS text \end{tabular}& \begin{tabular}[c]{@{}l@{}}Find if models can distinguish between \\ all the languages in a CS input \\ at the token level.\end{tabular} \\ \hline
Syntax & \begin{tabular}[c]{@{}l@{}}Dependency parse from \\ structural probe \end{tabular} & \begin{tabular}[c]{@{}l@{}}Train a structural probe to extract the dependency \\parse of sentences in English and Spanish. \\ The probe is used on CS data and the translations.\end{tabular} & \begin{tabular}[c]{@{}l@{}}Study the structures of CS input and \\ compare them with the structure \\ of the monolingual translations.\end{tabular} \\ \hline
Semantics & Semantic Text Similarity (STS) & \begin{tabular}[c]{@{}l@{}}Fine-tune PLMs on STS task in Spanish and English, \\which assigns a score on the similarity of two texts. \\ Use CS data and Spanish and English data to get \\ scores on different language pairs and sentence pairs.\end{tabular} & \begin{tabular}[c]{@{}l@{}} Determine whether PLMs are consistent\\  in encoding meaning of CS text \\compared to monolingual representations. \end{tabular} \\ \hline
\end{tabular}%
}

\caption{Summary of tested dimensions and associated conducted experiments.}
\label{tab:experiments}
\end{table*}

\section{Detection}
\label{sec:detection}

We use probes to conduct a layer-wise exploration of the PLMs in order to find if models are able to differentiate between monolingual and CS input.
These experiments fall under (1) sentence classification, in which we train probe classifiers to differentiate between monolingual and CS sentences, and (2) an LID task, in which a probe is trained to detect the natural language of a token, given a CS sentence.
These experiments are designed to understand whether PLMs have access to source language information in processing CS data, and if so, we wish to determine if the information pertaining to language varies between the layers of different language models.

\subsection{Methods}
\label{sec:detection-methods}
 For the experiments dealing with sentence and token classification (LID), we use the following CS datasets: SentiMix 2020~\citetlanguageresource{SentiMix}, and  CALCS 2016 LID dataset~\citetlanguageresource{LID}.
Additionally, we use monolingual datasets in Spanish and English created for the ProfNER 2021 shared task~\citetlanguageresource{miranda-escalada-etal-2021-profner}.
The ProfNER dataset consists of tweets in Spanish and English. 
We use the ProfNER data together with the SentiMix data to create a balanced dataset containing text in Spanish (es) (4,000 examples with label 0), English (en) (4,000 examples with label 0), and CS (8,000 examples with label 1). 
This way we are sure to have balanced classes for training the probe classifier. 
We use an 80-10-10 split to train, validate and test the classifier.
This combined dataset is used on the\textit{ sentence classification} task, in which we train a probe classifier to distinguish between monolingual and code-switched sentences.

For the LID \textit{token classification task}, we use two datasets, CALCS 2016 LID dataset, and SentiMix 2020, which contain language ID tags for each token.
The possible labels for the LID task are \textit{lang1} (en), \textit{lang2} (es), \textit{other}, \textit{ne} (named entities), \textit{fw} (a language different from \textit{lang1} and \textit{lang2}), \textit{mixed} (partially in both languages), \textit{unk} (unrecognizable words), \textit{ambiguous} (either one language or another)~\cite{Lince2020}.
For the LID task, we train probes separately on the datasets to see how probe performance changed, if at all. 
In the CALCS 2016 dataset, 7,986 examples contain both source languages in the same sentence. 
There are 21,030 train examples in this dataset, meaning that 38\% of the data the probe was trained and tested with contained true instances of intra-sentential CS.
The SentiMix dataset, on the other hand, contains 11,783 intra-sentential CS examples. 
In total, 96\% of the examples used to train, validate and test probes on the SentiMix dataset contain instances of CS.
For both the sentence and token classification tasks, we report the average F1 score across 5 seeds for each layer and model. 
For each probe, we use a batch size of 32, and learning rate of 1e-3.

\subsection{Results}
\label{sec:detection-results}

\begin{figure}[t]
     \centering
     \includegraphics[scale=0.3]{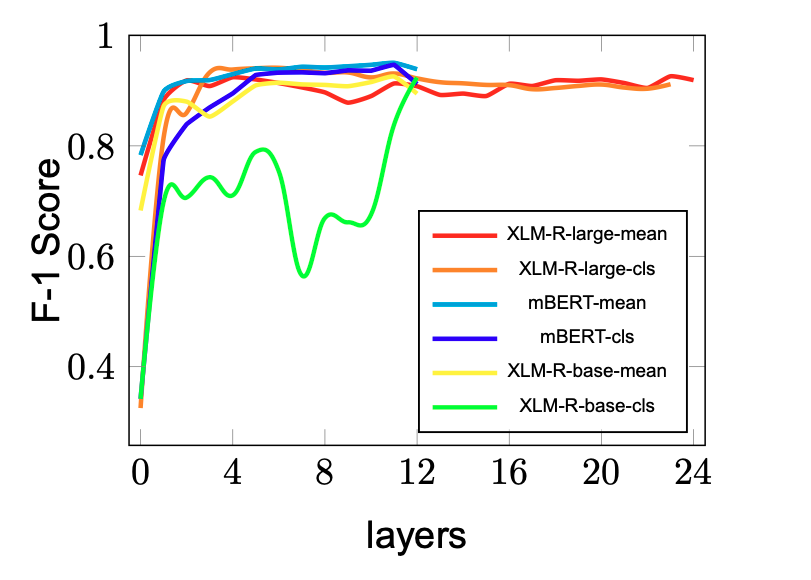}
    \caption{Mean F-1 Scores across layers for the sentence classification task for each of the PLMs studied. In this task, probe classifiers learn to distinguish between CS and monolingual text.} 
    \label{fig-sequence-classification}
 \end{figure}

\begin{figure}[t]
 \centering
 \includegraphics[scale=0.3]{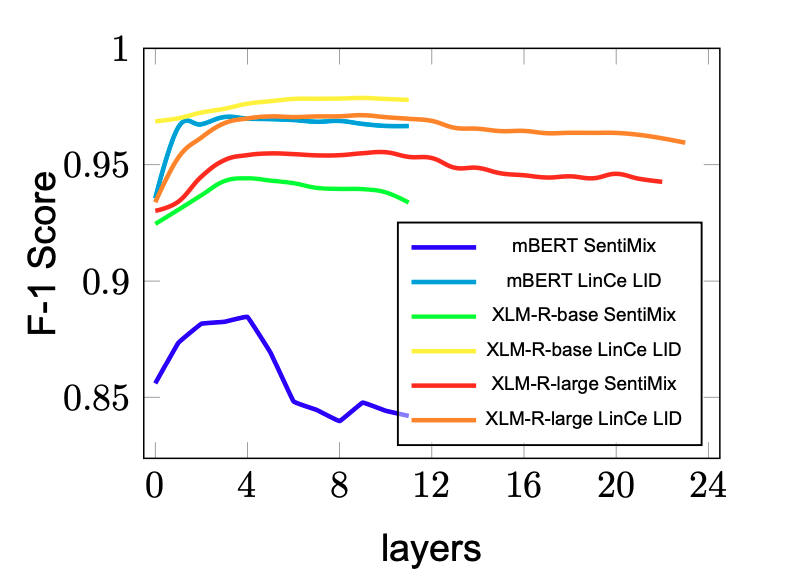}
    \caption{LID model mean F-1 Scores across layers for the probe classifiers. 
    In this task, probe classifiers learn the LID of the tokens in CS sentences. 
    }
    \label{fig:lid}
\end{figure}

The results of the detection experiments are displayed in figures~\ref{fig-sequence-classification} and~\ref{fig:lid}.
The probe results indicate that PLMs are, in general, able to distinguish between CS text and monolingual text.
Interestingly, for the sentence classification task, CLS pooling for XLM-R-base causes the probe to struggle with the sentence classification task, although by the latter layers (10, 11) the F1 score begins to match that of the other probe classifiers corresponding to those layers. 
This could be because the CLS token may not fully capture information relating to the differences in multiple languages, while mean pooling considers the entire input sequence, thereby capturing the differences in languages better.
On the other hand, we can see that XLM-R-large CLS pooling is effective for the task, indicating, perhaps, that models with more parameters are able to encode this information in the CLS token.
Overall, it seems that for the base models, the mean pooling strategy is more effective than using the CLS token, likely because mean pooling allows us to consider the full input sequence.

For the LID task, our results indicate that PLMs seem to have language information at the token level embedded within them from early layers in the models, see figure \ref{fig:lid}.
This indicates that PLMs may have encoded knowledge on features such as vocabulary or morphology for different languages.
Given that the probe classifiers achieve high F-1 scores for both datasets, SentiMix2020 and CALCS 2016, it may be the case that this information is used throughout all layers.
In our experiments, mBERT seems to struggle when compared to the other models on the SentiMix dataset.
This could be due to a combination of two things: (1) XLM-RoBERTa generally outperforms mBERT on cross-lingual classification~\cite{Conneau2018}, and (2) the SentiMix dataset may be more challenging than CALCS 2016 because SentiMix contains more CS examples. 
Regardless, the average F1 score for mBERT on the SentiMix dataset remains at 0.84 and above, indicating that the model still has some information at the token level to do well at the LID task. 
Generally, the probe classifiers trained and tested on the SentiMix dataset exhibit a drop in performance in contrast with the probes trained on the CALCS 2016 dataset. 
This likely due to the amount of CS examples in each of the datasets~\ref{sec:detection-methods}, which may mean that the SentiMix dataset may be more representative of the LID task for CS text.
Overall, these experiments show that PLMs are very effective at detecting CS text at both the sentence and token levels, even with a more challenging dataset.

\section{Syntax}
\label{sec:syntax}

To evaluate the effective generalisability of the inferred structure of CS data, we evaluate the extent to which CS data is similar to the majority language in that text. 
We do this using our dataset \textbf{\textit{r-CS}} which has translations into the source languages. 
We set this up using the structural probe which was developed by~\citet{Hewitt2019} to allow us to extract the structural information captured in CS text and evaluate if the structure is closer to one language compared to another.
We repeat this experiment with synthetic data to ensure that we have a more controlled way of measuring this. 
The results of these experiments are presented in Section~\ref{sec:syntax-results}. 

\subsection{Methods}
We use the structural probe developed by~\citet{Hewitt2019}, and use the code base by~\citet{Chi2020} to train
 a probe to recreate the dependency tree structure using Universal Dependencies (UD) datasets~\cite{Nivre2020}.
 We train the probe using mBERT on both UD Spanish ancora~\citetlanguageresource{ancora} and UD English EWT~\citetlanguageresource{silveira}, and validate the probe on the test partition of these datasets to ensure high performance on monolingual data, based on two evaluation metrics: Spearman correlation between predicted and true word pair distances, and on undirected, unlabelled attachment score (UUAS), the percentage of undirected edges placed correctly~\cite{Chi2020}.
 We train the structural probe on layer 7 of mBERT and use a maximum rank of 128 to recover the path length between each pair of words in a sentence~\cite{Hewitt2019}. 
We do this because of reasons found in Section~\ref{sec:lit-review}.
Once we have ensured that the probe recovers appropriate dependency parses for monolingual data, we extract dependency parses from the probe using the CS examples from our \textit{\textbf{r-CS}} dataset.  
We then generate dependency parses for the translations in English and Spanish from our dataset.

Due to the lack of gold labels for CS-dependency parses, we have decided to use the graph edit distance (GED) between the dependency parse of a code-mixed sentence and the dependency parse of the monolingual translations as given by the syntax probe. 
Using GED tells us how many changes a dependency parse need to undergo to resemble another dependency parse. 
Therefore, we analyse the distances between the GEDs of a code-mixed sentence and the translated monolingual sentences, with the aim of finding if the CS structure aligns more with one of the source languages when compared to the other. 
To do this, we use the NetworkX python library~\cite{Hagberg2008}.
Finding the GED between two graphs can often be slow for graphs containing more than 10 nodes~\cite{Hagberg2008}.
This is the reason we select a subset of the CS dependency parses derived from the structural probe, specifically, examples that contain 10 nodes or fewer, totalling 118 examples. 
We then extract the dependency parses for the translations of the 118 CS examples and compare the distances . 
We repeat these experiments with the synthetically generated CS data: \textit{\textbf{randCS}} and \textit{\textbf{NP-CS-es}} and \textbf{\textit{NP-CS-en}}~\ref{sec:datasets}.

\subsection{Results}
\label{sec:syntax-results}
Our results associated with syntax are presented in table~\ref{tab:spearman-syntax}.  
These results show that there is a strong correlation in the graph edit distances between real CS text and the monolingual translations of that text.
In order to assess the potential correlation between the distances for the different language pairs, we use Spearman correlation.
For example, to compare monolingual text to real instances of CS text, we use the\textbf{\textit{r-CS}} data and find the distance between those CS examples and the corresponding Spanish translations, then we do the same for the English translations.
We then find the Spearman statistic between these two sets of distances.
The results indicate that the model generates CS dependency parses that are similar in distance to the monolingual parses, that is, the dependency parses are not closer in distance to one language compared to another. 
This is the case despite our dataset containing more English tokens than Spanish tokens, see table~\ref{tab:cs_dataset_stats}.
The results also show that when synthetic CS examples are used, the correlation of distances between the CS examples and parallel translations drops, perhaps indicating that some of the synthetic CS examples lack syntactic structure.

\section{Semantics}
\label{sec:semantics}
One of our aims is to discover whether PLMs are able to effectively capture the meaning of code-mixed sequences. 
We carry out an intrinsic exploration to see how the representations of code-switched sentences compare with monolingual sentences.
We want to find if PLMs are \textit{consistent }in representing semantic information in CS text when compared to semantic representations of monolingual text. 
To do this, we fine-tune all the PLMs on the semantic text similarity (STS) task using monolingual benchmark STS data in Spanish and English.

\subsection{Methods}
PLMs do not generate semantically meaningful sentence embeddings unless specifically trained for this, therefore we must fine-tune the models on the STS task.
We build on work by~\citet{Madabushi2022} to set up the semantic experiments. 
~\citet{Madabushi2022} developed a method to find whether a PLM is consistent in scoring two sentences or expressions with similar meaning.
Given two input sentences, the models must return an STS score between 0 (least similar) and 1 (most similar).
We adopt this method to find if a model, after it is fine-tuned on the STS task, is consistent in scoring monolingual sentences and CS sentences. 
The PLMs fine-tuned on the STS task should be consistent in scoring monolingual sentences and CS sentences. 
That is,  the sentence similarities of ($i_{es}, j_{es}$) and ($i_{en}, j_{en}$), should approximate the similarities between ($i_{cs}, j_{cs}$) and  ($i_{es}, j_{es}$) and ($i_{en}, j_{en}$). 
We formalise this in Eq. ~\ref{eq:semantic-sim}.

\begin{equation}
    sim(S^{l_1}_i,S^{l_2}_j) = sim(S^{cs}_i,S^{l}_j)
    \label{eq:semantic-sim}
\end{equation}
where $sim$ represents the cosine similarity. $S$ is a sentence in the dataset. The languages of the sentences are encoded by $(l_1,l_2,l) \in \{es,en\}\times \{es,en\} \times \{es,en,cs\}$. 

The indexes represented by $(i,j)\in N^2$ correspond to the sentences in the dataset of length $N$. 

We use the dataset\textbf{ \textit{r-CS} }to get similarities between the language pairs listed in table~\ref{tab:semantic-results}.
The similarities output by the fine-tuned PLMs are compared to each other using Spearman Rank Correlation.
All PLMs were fine-tuned using a batch size of 8. 
The base models were fine-tuned using a learning rate of 2e-5, and XLM-R-large was fine-tuned with a learning rate of 2e-6. 

\begin{table}[]
\resizebox{\columnwidth}{!}{%
\begin{tabular}{|c|c|c|}
\hline
\textbf{lang-pair 1} & \textbf{lang-pair 2} & \textbf{Spearman statistic} \\ \hhline{|=|=|=|}
\cellcolor{black!15} cs vs. en               & \cellcolor{black!15}cs vs. es                & \cellcolor{black!15} 0.8308                      \\ \hline
NP-CS-en vs. en& NP-CS-en vs. es             & 0.6876                      \\ \hline
NP-CS-es vs. en             & NP-CS-es vs. es             & 0.7564                      \\ \hline
randCS vs. en           & randCS vs. es           &0.6983                     \\ \hline
\end{tabular}%
}
\caption{Spearman rank for correlation between distances of code-mix and monolingual text. 
Results on real CS data is highlighted.}
\label{tab:spearman-syntax}
\end{table}

\subsection{Results}
Our results associated to semantics are presented in tables~\ref{tab:semantic-results} and~\ref{tab:semantic-results-synthetic}. In general, these results show that the models are able to capture the meaning of naturally occurring code-mixed sentences in a way that aligns with how they capture the meanings in monolingual sentences. 
These results show that the strongest correlations are between $sim(cs_i, cs_j)$ - $sim(en_i, en_j)$ and between $sim(cs_i, en_j)$ - $sim(en_i, es_j)$.
In general, though, for all models, the Spearman rank statistic comparing all language pairs is high, meaning that the PLMs, fine-tuned on monolingual data, have the capacity to effectively capture and represent semantic relationships between CS text and monolingual text in a manner consistent with how they represent those relationships between the monolingual pairs.
We also conduct these experiments using synthetic CS data,  \textit{\textbf{randCS}} and \textit{\textbf{NP-CS-es}} and \textbf{\textit{NP-CS-en}}; table~\ref{tab:semantic-results-synthetic} contain the results for the experiments with the synthetic data.
These results may indicate that the model is not able to capture meaning consistent to the monolingual translations of these examples.
This may be for a number of reasons, perhaps because these generations are not guaranteed to be well-formed CS text, it may indicate that the model relies on the syntactic structure of a sentence to provide semantic similarity.
Further experiments with different types of synthetically generated CS would be needed for proper analysis.

\begin{table}[]
\resizebox{\columnwidth}{!}{%
\begin{tabular}{|c|c|ccc|}
\hline
\multirow{2}{*}{\textbf{l-pair-1}} & \multirow{2}{*}{\textbf{l-pair-2}} & \multicolumn{3}{c|}{\textbf{cosine spearman}}\\ 
    & & mBERT& XLM-R-base& XLM-R-large \\ \hhline{|=|=|===|}
en-en           &   cs-cs           & 0.8503& 0.8208& 0.8256                                                                              \\ \hline
es-es             &     cs-cs         & 0.7892& 0.7655& 0.7799                                                                              \\ \hline
en-es             & cs-en             & 0.8695& 0.8656& 0.8704                                                                              \\ \hline
en-es             & cs-es             & 0.7266& 0.6947& 0.7200                                                                              \\ \hline
\end{tabular}%
}
\caption{Spearman rank statistic for the cosine similarity between language pair 1 (l-pair-1) and language pair 2 (l-pair-2).}
\label{tab:semantic-results}
\end{table}

\begin{table}[]
\resizebox{\columnwidth}{!}{%
\begin{tabular}{|c|c|ccc|} \hline 
\multirow{2}{*}{\textbf{l-pair-1}} & \multirow{2}{*}{\textbf{l-pair-2}} & \multicolumn{3}{c|}{\textbf{cosine spearman}}\\
    & & mBERT& XLM-R-base& XLM-R-large \\ \hhline{|=|=|===|}
randCS-randCS& en-en             & 0.0106& -0.0028& 0.0105\\ \hline  
randCS-randCS& es-es             & 0.0091& 0.0177& 0.0189\\ \hline  
 NPesCS-NPesCS& en-en             &0.0027& 0.0009& -0.0029\\ \hline  
 NPesCS-NPesCS& es-es             &0.0188& 0.0208& 0.0021\\ \hline  
 NPenCS-NPenCS& en-en             &0.0151& 0.0009& 0.0048\\ \hline  
 NPenCS-NPenCS& es-es             &0.0188& 0.0205& 0.0065\\ \hline 
en-es             & randCS-en& 0.0184& 0.0114& 0.0212\\ \hline  
en-es             & randCS-es& 0.0043& 0.0154& 0.0156\\ \hline  
 en-es             & NPesCS-en&0.0102& 0.0030& 0.0130\\ \hline  
 en-es             & NPesCS-es&0.0011& 0.0223& 0.0108\\ \hline  
 en-es             & NPenCS-en&0.0107& 0.0111& 0.0168\\ \hline  
 en-es             & NPenCS-es&0.0056& 0.0221& 0.0152\\ \hline 
\end{tabular}%
}
\caption{Semantic experiments results with synthetic CS data and the original monolingual translations of the r-CS-syn dataset.}
\label{tab:semantic-results-synthetic}
\end{table}

\section{Discussion}
\label{sec:discussion}
The results across all categories of experiments seem to indicate that PLMs are likely to have the potential to generalise to being able to handle CS text.
We find that PLMs are effective at detecting CS text at a sentence level and token level in our detection experiment.
We find that dependency parses generated by the model are not more similar in distance to one language or another in our syntax experiments that is, experimental results reveal a strong correlation in the distances of dependency parses between English (cs-en) and Spanish (cs-es).
We find as well that the models are consistent in capturing meaning representations of real CS text, but are unable to do so for synthetically generated text using our generation methods. 
They seem to capture syntactic structure and semantic meaning across real CS text, without being trained on CS text, see tables~\ref{tab:spearman-syntax} and~\ref{tab:semantic-results}.

Our findings show that PLMs are able to generalise across CS text containing Spanish-English language pair.
They also show that in general, performance of the probes degrades when using synthetic CS text. 
In the syntax experiments, the correlation between the distances diminish, though this could be attributed to the difference in distribution of a synthetically generated CS example when compared to a well-formed CS example. 
Although we used methods found in literature to generate the synthetic examples, there is no guarantee of these methods producing a naturalistic CS sentence. 
In the semantic experiments, the Spearman correlation statistic drops to nearly zero across all models when using synthetically generated CS text, which may be due to the loss of grammatical correctness.
This may show that PLMs rely on the syntactic structure of a sentence to provide the semantics.
Further experiments with different types of synthetically generated CS would be needed for a proper analysis.

In general, the experimental results across detection, syntax and semantics, show that for real CS text containing languages that are closely related, such as Spanish and English, PLMs may contain enough linguistic information from the source languages to handle the mixed language text.
This is promising, because if monolingual data can be harnessed for some tasks, then the scarcity of data in certain CS language pairs can be mitigated by the PLMs ability to generalise.
We would like to explore this idea in future research.

\section{Conclusion and Future Work}
\label{sec:conclusion}
In this paper, we present our finding on how pre-trained models handle code-switched text.
Our contributions include a novel dataset of CS text and translations into the source languages, Spanish and English.
Additionally, we extend probing work to code-switching in Spanglish in the areas of syntax and semantics.
We carry out experiments in detection, syntax and semantics, to explore how PLMs capture CS text.
We find that PLMs seem to be effective at detecting CS text.
In the future, we hope to explore PLMs abilities to learn from monolingual data for use on CS text, experiment with further synthetic data generation methods, and to expand to other languages.

\section{Acknowledgements}
The computations described in this research were performed using the Baskerville Tier 2 HPC service (\url{https://www.baskerville.ac.uk/}). Baskerville was funded by the EPSRC and UKRI through the World Class Labs scheme (EP/T022221/1) and the Digital Research Infrastructure programme (EP/W032244/1) and is operated by Advanced Research Computing at the University of Birmingham.

\section{Optional Supplementary Materials}

\subsection{Limitations}
Our work only explores how models embed code-switched data for Spanglish, although this was done so as not to confound model capabilities by other aspects of language use, in the future, we would like to extend our explorations to languages such as Hinglish.
Doing so will allow us to see the extent to which PLMs generalise to different language pairs, especially pairs that are not closely related languages.
Our work only explores auto-encoder models, such as mBERT and XLM-RoBERTa, which does not offer a comprehensive view of how different types of models encode CS-text. In the future, we would like to explore the capabilities and degree to which models such as GPT encode CS text.

\subsection{Ethics Statement}
We do not use any private data, all data used is publicly available, or will become available after the end of the anonymity period.
The dataset that we create is collected from social media and may contain profanity or toxic content.
We work with one language pair for code-switching, out of many, and hope in the future to expand this to other CS language pairs, especially low-resource pairs.

\nocite{*}
\section{Bibliographical References}
\label{sec:reference}

\bibliographystyle{lrec-coling2024-natbib}
\bibliography{main}

\section{Language Resource References}
\label{lr:ref}

\bibliographystylelanguageresource{lrec-coling2024-natbib}\bibliographylanguageresource{LanguageResources}

\end{document}